\documentclass{bmvc2k}

\title{Energy-Based Residual Latent Transport for Unsupervised Point Cloud Completion}

\addauthor{Ruikai Cui}{ruikai.cui@anu.edu.au}{1}
\addauthor{Shi Qiu}{shi.qiu@anu.edu.au}{1,2}
\addauthor{Saeed Anwar}{saeed.anwar@csiro.au}{1,2}
\addauthor{Jing Zhang}{jing.zhang@anu.edu.au}{1}
\addauthor{Nick Barnes}{nick.barnes@anu.edu.au}{1}

\addinstitution{
 Australian National University,\\
 Canberra, ACT, AU
}
\addinstitution{
 Data61, CSIRO\\
 Canberra, ACT, AU
}

\runninghead{Student, Prof, Collaborator}{BMVC Author Guidelines}

\usepackage[utf8]{inputenc} % allow utf-8 input
\usepackage[T1]{fontenc}    % use 8-bit T1 fonts
\usepackage{hyperref}       % hyperlinks
\usepackage{url}            % simple URL typesetting
\usepackage{booktabs}       % professional-quality tables
\usepackage{amsfonts}       % blackboard math symbols
\usepackage{nicefrac}       % compact symbols for 1/2, etc.
\usepackage{microtype}      % microtypography
\usepackage{amsmath, amsthm, amssymb, amsfonts}
\usepackage{mathtools}
\usepackage{xfrac}
\usepackage{graphicx}
\usepackage{color, colortbl}
\usepackage{multirow}
\usepackage{tabularx}
\usepackage[ruled,vlined,noline]{algorithm2e}
\usepackage[normalem]{ulem}
\usepackage{csquotes}

\def\eg{\emph{e.g}\bmvaOneDot}

\SetKwInput{KwInput}{Input}                % Set the Input
\SetKwInput{KwOutput}{Output}              % set the Output

\begin{document}

\maketitle

\begin{abstract}
Unsupervised point cloud completion aims to infer the whole geometry of a partial object observation without requiring partial-complete correspondence. Differing from existing deterministic approaches, we advocate generative modeling based unsupervised point cloud completion to explore the missing correspondence. Specifically, we propose a novel framework that performs completion by transforming a partial shape encoding into a complete one using a latent transport module, and it is designed as a latent-space energy-based model (EBM) in an encoder-decoder architecture, aiming to learn a probability distribution conditioned on the partial shape encoding. To train the latent code transport module and the encoder-decoder network jointly, we introduce a residual sampling strategy, where the residual captures the domain gap between partial and complete shape latent spaces. As a generative model-based framework, our method can produce uncertainty maps consistent with human perception, leading to explainable unsupervised point cloud completion. We experimentally show that the proposed method produces high-fidelity completion results, outperforming state-of-the-art models by a significant margin.
\end{abstract}

%-------------------------------------------------------------------------
\section{Introduction}
\label{sec:intro}
Point cloud data is a fundamental representation of 3D geometry, contributing to numerous applications in robotics, auto-navigation, augmented reality, \emph{etc}.
Limited by viewing angle, occlusion, and acquisition resolution, raw point clouds are generally sparse and incomplete.
We argue that the completion of partial scans is not only essential for a better understanding of 3D scenes, but also beneficial for many downstream 3D computer vision tasks, including classification~\cite{pointnet, qiu2022geometric}, segmentation~\cite{GuoSurvey2021, qiu2021semantic}, and detection~\cite{Qi_2019_ICCV, qiu2021investigating}.
To this end, increasing attention has been dedicated to point cloud completion.

Pioneered by PCN~\cite{8491026}, supervised methods~\cite{9156338, Pan_2021_CVPR, Yu_2021_ICCV, Xiang_2021_ICCV} have achieved impressive completion results.
However, they usually rely on large-scale datasets with both partial and corresponding ground truth complete shapes, where the latter is hard to collect.
% in the real world.
To tackle the difficulty in data collection, unsupervised point cloud completion has recently gained popularity due to its capability of utilizing both synthetic and real-world datasets.
In the unsupervised setting, only unpaired samples from the partial observation domain and complete shape domain are provided, so a model needs to infer the partial-to-complete relationship.
Existing methods~\cite{chen2020pcl2pcl, wen2021c4c, Cai_2022_CVPR} address the problem mainly by mapping a partial shape to a latent code that can be decoded as a valid complete shape.
Nevertheless, a deterministic one-to-one mapping is assumed in existing methods, which can be biased because a partial shape can correspond to multiple complete shapes due to the missing geometries.

We introduce a generative model-based strategy to explore the one-to-many mapping inherent in point cloud completion;
thus, we can capture the prediction uncertainties, representing our ignorance about the complete shape space.
Specifically, we leverage an encoder-decoder architecture, where parameters of the encoder and decoder for complete and incomplete point clouds, respectively, are shared.
For an incomplete shape latent code, we assume it locates somewhere near its corresponding complete shape codes in the latent space.
%% ebm prior distribution discussion
Consequently, we design an energy-based model (EBM)~\cite{ackley1985learning} and deploy it in the latent space of the encoder-decoder architecture.
The latent-space EBM aims to learn a conditional distribution of complete shape code given a partial shape code.
By sampling with the gradient-based Markov chain Monte Carlo (MCMC) (\eg~Langevin dynamics~\cite{NEURIPS2019_378a063b}) initialized by a partial shape code, a latent code corresponding to a valid complete shape can be generated.

The vanilla Langevin dynamics~\cite{NEURIPS2019_378a063b} do not facilitate back-propagation since a computationally expensive second-order gradient is needed when back-propagating through the Langevin iterations.
Therefore, existing EBMs are either integrated into a deep neural network as a prior model~\citep{zhang2021learning, latent_ebm} or utilized with a pre-trained network~\cite{Zhao_2021_CVPR}.
Unlike existing techniques, we propose a residual sampling strategy that, for the first time, enables joint training of a latent-space EBM and a task-related generator.
Particularly, instead of sampling the conditional latent code of a complete shape directly,
we generate a residual that captures the gap for transporting a code from one domain (partial shape space) to the other (complete shape space).
By adding the residual back to the partial latent code, we achieve both complete latent code sampling, and gradient back-propagation with parameter updating for the encoder-decoder.

Our main contributions are summarized below:
\begin{itemize}
    \item We propose a novel energy-based latent transport mechanism, enabling generative modeling of the unsupervised point cloud completion task for the first time.
    \item We present a residual sampling strategy that allows joint training of a latent-space EBM and an encoder-decoder.
    \item Experimental results indicate that our model not only achieves state-of-the-art performance on synthetic (ShapeNet~\cite{shapenet2015}) and real-world (KITTI~\cite{Geiger2013IJRR}, ScanNet~\cite{dai2017scannet}, MatterPort3D~\cite{Matterport3D}) datasets, but is also capable of generating explainable uncertainty maps.
\end{itemize}

\section{Related Works}
\label{sec:related}

\textbf{Unsupervised Point Cloud Completion.}
As a pioneering work for unsupervised point cloud completion~\cite{chen2020pcl2pcl, zhang2021unsupervised, wen2021c4c, Cai_2022_CVPR, DBLP:journals/corr/abs-2111-11976}, Pcl2Pcl~\cite{chen2020pcl2pcl} proposed an adversarial learning-based approach, where they firstly train two autoencoders for incomplete and complete point clouds, respectively, then a generator~\cite{gan_ian} was used to transform latent code of the incomplete shape to that of the complete shape.
Following~\cite{chen2020pcl2pcl}, Cycle4Completion~\cite{wen2021c4c} introduced two cycle transformations between the latent space of the complete and partial shapes, achieving dual-direction transformation of the two shape codes.
However, it imposes a strong \enquote{one-to-one deterministic mapping} constraint~\cite{Zhu_2017_ICCV}, neglecting the uncertainty in the partial and complete shape domains.
To address this issue, ShapeInversion~\cite{zhang2021unsupervised} searched for a latent code in the latent space of a pre-trained GAN~\cite{gan_ian} with a partial-complete matching loss.
Nevertheless, the search, as a minimization program, is highly non-convex, where the optimization can be easily stuck in a local minimum, leading to poor completion quality.
More recently, Cai~\emph{et al.}~\cite{Cai_2022_CVPR} encoded a series of related partial point clouds into a unified latent space as a complete shape code and multiple occlusion codes, but a deterministic one-to-one mapping is assumed to perform completion, while a partial shape can have multiple valid, complete predictions due to occlusion.
Different from existing methods, we propose a novel energy-based latent transport module, with which we aim to model the distribution gap between the partial and the complete shape codes.
Further, as a generative model, our method can produce uncertainty maps representing ignorance of the model towards its prediction.

\noindent
\textbf{Energy-Based Models.}
Learning a data distribution for new sample generation is an important task in machine learning.
Recent works have shown that energy-based models (EBMs)~\cite{ackley1985learning} have a strong capability in modeling high dimensional data, such as images~\cite{NEURIPS2019_378a063b}, videos~\cite{8798892}, and point clouds~\cite{Xie_2021_CVPR}.
Yet, the capability of EBMs in point cloud completion remains unclear, which makes our method a research pioneer on the topic.
Despite leveraging EBMs in data space, Pang~\emph{et al}.~\cite{latent_ebm} proposed to jointly learn a latent space energy-based prior with a top-down generator network.
Our method differs from \cite{latent_ebm} as our EBM takes an output from a learnable module.
Therefore, gradient back-propagation is necessary to train the network while \cite{latent_ebm} does not support it.
LETIT~\cite{Zhao_2021_CVPR} is the most related art to our method, where an EBM is deployed in the latent space of a pre-trained variational autoencoder (VAE)~\cite{vae_kingma} to transform a latent code from one domain to another domain.
Despite that, the VAE and the EBM are trained in two stages since sampling from an EBM blocks gradient back-propagation to the encoder.
To the best of our knowledge, we achieve simultaneous training of a latent space EBM for the first time, thanks to the proposed residual sampling strategy.

\section{Method}
\label{sec:method}
\begin{figure}
  \centering
  \includegraphics[width=0.75\textwidth]{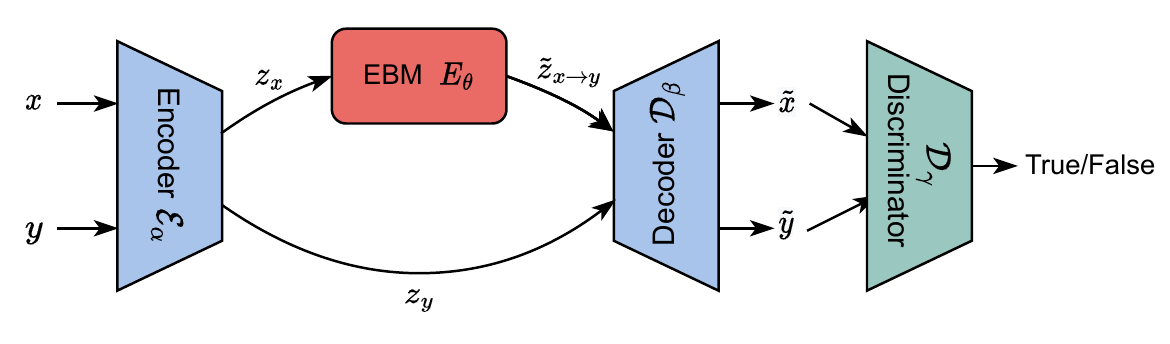}
  \caption{Model overview.
  Our model consists of 1) an encoder $\mathcal{E}_{\alpha}$ and a decoder $\mathcal{D}_{\beta}$ for both domain $\mathcal{X}$ and $\mathcal{Y}$,
  2) an EBM ($E_\theta$) that transports a partial shape latent code $z_x$ to a complete shape latent code
%   encoding
  $\tilde{z}_{x \rightarrow y}$,
  3) and a point domain discriminator $D_\gamma$.}
  \label{fig:framework}
\end{figure}

Let $x \in \mathcal{X}$ and $y \in \mathcal{Y}$ be samples from the partial and complete point cloud domain, respectively.
We are provided with samples from two marginal distributions $p(x)$ and $p(y)$ instead of the joint distribution $p(x, y)$ in the unsupervised point cloud completion task.
The goal of this task is to estimate the conditional distribution $p(y|x)$ with a point cloud completion model $p(y_{\mathcal{X}\rightarrow \mathcal{Y}}| x)$, where $y_{\mathcal{X}\rightarrow \mathcal{Y}}$ is a sample produced by translating a sample from the partial point cloud domain $\mathcal{X}$ to the complete point cloud domain $\mathcal{Y}$. Given the one-to-many mapping attribute inherent in the point cloud completion task, we present a generative unsupervised point cloud completion model, where a latent variable is introduced to explore the missing correspondence knowledge from the partial observation.
Our model consists of four main parts, namely
\textbf{1)} an encoder ($\mathcal{E}_\alpha$) for point cloud code extraction,
\textbf{2)} a latent-space energy-based model with energy function $E_\theta$ to fill the code gap between the partial point cloud and its corresponding completion,
\textbf{3)} a decoder ($\mathcal{D}_\beta$) for point cloud reconstruction
and \textbf{4)} a point domain discriminator ($D_\gamma$) to achieve adversarial learning (see Figure~\ref{fig:framework}).

Notably, the encoder projects a point cloud into the latent space, leading to a partial point cloud code $z_x$ or a complete point cloud code $z_y$.
Since $z_y$ is the representation of a shape with full geometry, we feed it directly to the decoder for point cloud reconstruction.
However, the decoder cannot effectively reconstruct the corresponding complete point cloud with the latent code $z_x$ of the incomplete shape due to information loss caused by partial observation.
The latent-space EBM is then introduced to transport $z_x$ to its corresponding complete shape latent code $\tilde{z}_{x\rightarrow y}$.
In this way, the same decoder can complete a partial observation with the inferred latent code $\tilde{z}_{x\rightarrow y}$.
Finally, adversarial training is presented to guarantee the quality of the inferred complete shape $\tilde{x}$ for effective unsupervised point cloud completion.

In the following sections, first, we introduce the energy-based transport module via residual sampling (Section~\ref{subsec:ebm_transport}).
Then, we discuss our learning pipeline to achieve joint training of our encoder-decoder framework (Section~\ref{subsec:joint_train_ebm_endecoder}).
Finally, we present our objective functions (Section~\ref{subsec:training_objective}) for the proposed generative unsupervised point cloud completion task.

\subsection{Energy-based Residual Latent Transport}
\label{subsec:ebm_transport}

\noindent\textbf{EBM for supervised point cloud completion - A preliminary:}
Following Pang \emph{et al.}~\cite{latent_ebm}, we define the conditional distribution of the complete shape latent code as: $p_\theta(z_y|z_x)=\frac{p_\theta(z_y,z_x)}{\int_\theta p_\theta(z_y,z_x)d z_y}=\frac{\exp[-E_\theta(z_{y},z_x)]}{Z(z_x;\theta)}$,
where $E_\theta(z_{y},z_x)$ is the energy function parameterized by a deep neural network, mapping the code pair $(z_y,z_x)$ to a scalar that measures their compatibility.
$Z(z_x;\theta)=\int\exp[-E_\theta(z_y,z_x)]dz_y$ is the normalizing constant.
For supervised point cloud completion, $E_\theta(z_{y},z_x)$ can be easily modeled as we have paired training samples.
In this case, given $z_x$, $z_y$ can then be achieved by Langevin dynamics~\cite{NEURIPS2019_378a063b} following:
\begin{equation}
    \label{paired_langevin_sampling}
    z_y^{t+1}=z_y^t-\frac{\delta^2}{2}\frac{\partial}{\partial z_y}E_\theta(z_y^t,z_x)+\delta\epsilon^t,
\end{equation}
where $t$ represents Langevin steps, $\delta$ is the step size, $\epsilon^t\sim\mathcal{N}(0,\mathbf{I})$ is the Gaussian random noise.
The prediction process via Langevin sampling in Eq.~\ref{paired_langevin_sampling} can be considered as finding $z_y$ to minimize the cost $E_\theta(z_y,z_x)$ given $z_x$.

% As discussed, for the unsupervised point cloud completion task, we have to access to the joint distribution of $p(x,y)$, thus we cannot directly design the energy function $E_\theta(z_y,z_x)$. Further, we claim that the code gap between the partial observation and it's corresponding completion should not be significant. In this case, we aim to learn the code gap or the residual of the latent code given only $z_x$ as input.

\noindent\textbf{Challenges and solution of applying EBM for unsupervised point cloud completion:}
In the unsupervised setting, we have no access to paired partial-complete point cloud samples, thus, we are incapable of directly modeling the compatibility of $z_x$ and $z_y$ through the energy function $E_\theta(z_y,z_x)$.
Alternatively, we model the conditional distribution of the residual $r_{xy}$ instead, representing the distribution gap between the latent space of the two domains as:
\begin{equation}
    \label{residual_ebm}
    p_\theta(r_{xy}|z_x)=\frac{p_\theta(r_{xy},z_x)}{\int_\theta p_\theta(r_{xy},z_x)d z_x}=\frac{\exp[-E_\theta(r_{xy},z_x)]}{Z(r_{xy};\theta)}.
\end{equation}

Different from $E_\theta$ for the conditional distribution of complete code in the supervised setting, the energy function $E_\theta(r_{xy},z_x)$ in Eq.~\ref{residual_ebm} is designed to build the bridge between $z_x$ and $z_y$, which we term \enquote{latent code transport}.
The benefits of learning the residual instead of the complete code mainly lie in
(see Section~\ref{subsec:joint_train_ebm_endecoder}):
\textbf{1)} No paired partial-complete point cloud data is needed for the residual-based EBM, leading to unsupervised point cloud completion.
Particularly, we can define the initial state of $r_{xy}$ as 0, representing no domain gap between partial-complete point cloud codes.
The model is trained to gradually refine $r_{xy}$.
\textbf{2)} It leads to joint training of the EBM and the encoder-decoder framework, which is significantly different from existing techniques, where the two modules are updated separately \cite{latent_ebm,zhang2021learning}.

\subsection{Joint Training of EBM and the Encoder-Decoder Framework}
\label{subsec:joint_train_ebm_endecoder}
For the conditional distribution $p_\theta(r_{xy}|z_x)$ modeled with an energy-based model, we can sample the residual $r_{xy}$ given a partial shape code $z_x$, by running several Langevin steps similar to Eq.~\ref{paired_langevin_sampling}.
% via:
% \begin{equation}
%     \label{unpaired_langevin_sampling}
%     r_{xy}^{t+1}=r_{xy}^t-\frac{\delta^2}{2}\frac{\partial}{\partial r_{xy}}E_\theta(r_{xy}^t,z_x)+\delta\epsilon^t,
% \end{equation}
% where $t$, $\delta$ and $\epsilon^t$ are defined the same as in Eq.~\ref{paired_langevin_sampling}.
% \SA{The reader knows this as it is the continuation of the previous equation. It is better to remove it from here}
With multiple ($K$) Langevin steps, it has been proved that the final $r_{xy}=r_{xy}^K$ is sampled from the true conditional residual distribution $q(r_{xy}|z_x)$~\cite{latent_ebm}.
In this way, a complete shape code $\tilde{z}_{x \rightarrow y}$ can be obtained as:
\begin{equation}
    \label{eq:add_residual}
    \tilde{z}_{x \rightarrow y} = z_{x} + \Omega({r_{xy}})
\end{equation}
Here, we apply the stop gradient operation ($\Omega(\cdot)$)~\cite{NEURIPS2019_378a063b} to avoid unfolding the Langevin dynamics iteration and involving the second-order gradient of $r_{xy}$ in future gradient computation, which is computationally expensive as shown in our supplementary material.
Specifically, we define the parameter set of the encoder as $\alpha$ and the loss function as $\mathcal{L}$.
The encoder can be trained with standard gradient descent via: $\frac{\partial \mathcal{L}}{\partial \alpha} =
    \frac{\partial \mathcal{L}}{\partial \tilde{z}_{x \rightarrow y}}
    \frac{\partial \tilde{z}_{x \rightarrow y}}{\partial z_{x}}
    \frac{\partial z_{x}}{\partial \alpha}$.
% \begin{equation}
%     \label{eq:encoder_gradient_unsimplied}
%     \frac{\partial \mathcal{L}}{\partial \alpha} =
%     \frac{\partial \mathcal{L}}{\partial \tilde{z}_{x \rightarrow y}}
%     \frac{\partial \tilde{z}_{x \rightarrow y}}{\partial z_{x}}
%     \frac{\partial z_{x}}{\partial \alpha}
% \end{equation}
With the stop gradient operator $\Omega(\cdot)$, the gradient
% Eq.~\ref{eq:encoder_gradient_unsimplied}
can be simplified as
$
\frac{\partial \mathcal{L}}{\partial \alpha} =
\frac{\partial \mathcal{L}}{\partial z_{x}}
\frac{\partial z_{x}}{\partial \alpha}
$, which can be supported by automatic differentiation.

% Following the conventional explicit generative model learning pipeline~\cite{larning_deep_ebm,a_theory_of_deep_gennet}, t
The EBM module can be trained via maximizing the log-likelihood~\cite{larning_deep_ebm,a_theory_of_deep_gennet} as: $L_{\text{ebm}} = \frac{1}{N} \sum_{i=1}^{N}{\log p_{\theta}(r_{xy}^i|z_x^i)}$, where $N$ is the number of the partial point clouds.
% shapes.
The
% learning
gradients can
% then
be obtained as: $\Delta\theta=\frac{1}{N}\sum_{i=1}^N\{\mathbb{E}_{p_\theta(r_{xy}|z_x^i)}\left[\frac{\partial}{\partial\theta}E_\theta(r_{xy},z_x^i)\right]-\frac{\partial}{\partial\theta}E_\theta(r_{xy}^i,z_x^i)\}$.
% Note that i
In practice, we feed $\tilde{z}_{x \rightarrow y}$ to
% the energy function
$E_\theta$, leading to: $\Delta\theta=\frac{1}{N}\sum_{i=1}^N\{\mathbb{E}_{\tilde{z}_{x \rightarrow y}\sim p_\theta}\left[\frac{\partial}{\partial\theta}E_\theta(\tilde{z}_{x \rightarrow y}^i)\right]-\frac{\partial}{\partial\theta}E_\theta(r_{xy}^i,z_x^i)\}$.

The first term of $\Delta\theta$ is tractable. However, we cannot directly estimate $\frac{\partial}{\partial\theta}E_\theta(r_{xy}^i,z_x^i)\}$ as the paired residual and partial code $(r_{xy},z_x)$ is unavailable. We find that $E_\theta(r_{xy}^i,z_x^i)$ is minimized when $r_{xy}$ can indeed represent the true partial-complete code gap. We then claim that the minimization of $E_\theta(r_{xy}^i,z_x^i)$ is equivalent to finding $z_y$ of the complete point cloud code that minimizes the same energy function $E_\theta$. In this way, we approximate $E_\theta(r_{xy},z_x)$ of the partial point cloud with $E_\theta(z_y)$ of the complete point cloud with:
\begin{equation}
    \label{eq:eqv_log_likelihood}
    \Delta\theta
    = \frac{1}{N}\sum_{i=1}^N{\mathbb{E}_{\tilde{z}_{x \rightarrow y} \sim p_{\theta}}[\frac{\partial}{\partial \theta} E_{\theta}(\tilde{z}_{x \rightarrow y}^i)] - \mathbb{E}_{z_{y} \sim p_{z_{y}}}[\frac{\partial}{\partial \theta}{E_\theta(z_y)}]} + C,
\end{equation}
where $C$ is a constant which can be ignored during training.
We use the proposed residual sampling strategy to sample a $\tilde{z}_{x\rightarrow y}$ from $p_{\theta}$, and Adam \citep{KingmaB14} with $\Delta{\theta}$ to update $\theta$.
Given the inferred latent code for both the partial and complete observation, the decoder $\mathcal{D}_{\beta}$ parameterized with $\beta$ can then be updated with stochastic gradient descent.
% \Jing{I'm here!}

\subsection{Training Objectives}
\label{subsec:training_objective}

\noindent \textbf{Reconstruction loss:}
We adopt Chamfer Distance (CD) as the reconstruction loss.
Given two sets of point clouds $S_1$ and $S_2$, CD measures the point cloud similarity via:
\begin{equation}
  \text{CD}(S_1, S_2) = \frac{1}{|S_1|} \sum_{p_1\in S_1}{\min_{p_2\in S_2}{\|p_1-p_2\|_2} }
  + \frac{1}{|S_2|} \sum_{p_2\in S_2}{\min_{p_1\in S_1}{\|p_2-p_1\|_2} },
  \label{eq:cd}
\end{equation}
where $p_1$ and $p_2$ are points from $S_1$ and $S_2$, respectively.
Given the reconstructed complete point cloud $\tilde{y}$ and the ground truth $y$, the reconstruction loss is defined as:
$
    \mathcal{L}_{\text{recon}} = CD(y, \tilde{y})
$

\noindent \textbf{Fidelity loss:}
\enquote{Fidelity loss} is used to measure how much geometrical details of a partial observation are preserved in the complete prediction.
Specifically, the fidelity loss is defined as a Unidirectional Chamfer Distance (UCD) as shown in Eq.~\ref{eq:fidelity}:
\begin{equation}
    \mathcal{L}_{\text{fidelity}} = \frac{1}{|x|}\sum_{p_1 \in x}{\min_{p_2 \in \tilde{x}}\|p_1 - p_2\|_2},
    \label{eq:fidelity}
\end{equation}
where $\tilde{x}$ is the reconstructed complete point cloud from the inferred complete code $\tilde{z}_{x \rightarrow y}$, and $x$ is a partial shape input.

\noindent \textbf{Adversarial loss:} We introduce adversarial training to our framework with a discriminator parameterized by a deep neural network to achieve regularizing point predictions to be complete shapes.
Specifically, we define adversarial loss following \cite{Geometric_gan,wgangp} as:
\begin{equation}
    \mathcal{L}_{\text{adv}} = \mathbb{E}[\min(0, -1 + D_\gamma(\tilde{y}))] + \mathbb{E}[\min(0, -1 - D_\gamma(\tilde{x}))]
\end{equation}
where $\mathcal{D}_{\gamma}$ is the discriminator, $\tilde{x}$ and $\tilde{y}$ are point clouds reconstructed given partial input and complete input, respectively.

\noindent \textbf{EBM loss:} The EBM parameters $\theta$ can be updated with $\Delta\theta$ in Eq.~\ref{eq:eqv_log_likelihood}.
Additionally, we follow~\cite{NEURIPS2019_378a063b} to add a weak $\ell_2$ normalization term on the energy magnitude for numerical stability.
Therefore, the loss function for the latent transport module is defined as:
\begin{equation}
    \mathcal{L}_{\text{ebm}} = -L_{\text{ebm}}
    + \lambda(E_{\theta}({z_y})^2
    + E_{\theta}(\tilde{z}_{x\rightarrow y})^2)
\end{equation}
where $\lambda$ is the weight for the regularization term, and we define $\lambda=0.1$ empirically.

Given the definition of the above four types of loss functions, we train the proposed model using the following steps:
\textbf{1)} For partial shape $x$ and complete shape $y$, the encoder generates their corresponding latent code $z_x$ and $z_y$;
\textbf{2)} The energy-based transport translates $z_x$ to its corresponding complete shape latent code $\tilde{z}_{x\rightarrow y}$, where the Langevin step size $\delta^2=0.05$;
\textbf{3)} The decoder reconstructs $\tilde{x}$ and $\tilde{y}$ from $\tilde{z}_{x\rightarrow y}$ and $z_y$, representing the reconstructed complete shape of $x$ and that of $y$;
\textbf{4)} The encoder-decoder framework with parameters $\{\alpha,\beta\}$ is updated with the loss function $\mathcal{L}_{\text{ed}}=\mathcal{L}_{\text{recon}}+\lambda_1\mathcal{L}_{\text{fidelity}} - \lambda_2 \mathbb{E}[\mathcal{D}_{\gamma}(\tilde{x})]$ where the last term ($-\mathbb{E}[\mathcal{D}_{\gamma}(\tilde{x})]$) is adopted from Miyato \emph{et al.}~\citep{miyato2018spectral} to regularize the distribution of predicted point clouds in the complete shape domain.
Empirically, we set $\lambda_1=2$ and $\lambda_2=1$;
\textbf{5)} The latent transport module is updated with $\mathcal{L}_{\text{ebm}}$;
\textbf{6)} The discriminator is updated with $\mathcal{L}_{\text{adv}}$.

During testing, given the partial point cloud $x$, we first obtain its latent code $z_x=\mathcal{E}_\alpha(x)$.
Then we feed it to the latent transport module with an initial state of $r_{xy}^0=0$.
We run $K=8$ Langevin steps
% via Eq.~\ref{unpaired_langevin_sampling}
to obtain the residual $r_{xy}=r_{xy}^K$, which is used to generate the complete shape code $\tilde{z}_{x\rightarrow y}$ via Eq.~\ref{eq:add_residual}.
Finally, we generate the complete shape $\tilde{x}=\mathcal{D}_\beta(\tilde{z}_{x\rightarrow y})$.

\section{Experiments}
\label{sec:exp}

\subsection{Implementation Details}

We implement the encoder with the PointNet++~\cite{NIPS2017_d8bf84be} set abstraction operation and point transformers~\cite{Zhao_2021_ICCV}.
There are three set abstraction layers in our encoder; following each is a point transformer block.
The decoder consists of a latent code projection layer, three self-attention blocks~\cite{NIPS2017_3f5ee243}, and a multi-layer perceptron~\cite{haykin1994neural} with one hidden layer.
The discriminator is implemented in the same way as PU-GAN~\cite{li2019pugan}.

\subsection{Performance on ShapeNet Dataset}
\textbf{Dataset}.
In the footstep of \cite{wen2021c4c, chen2020pcl2pcl}, we evaluate our method on the 3D-EPN dataset~\cite{dai2017complete}, which is a point cloud completion benchmark derived from the ShapeNet~\cite{shapenet2015} dataset.
Eight incomplete shapes are produced for each 3D shape by projecting a 3D shape to a 2.5D depth image with one of eight fixed viewpoints and then back-projected to 3D coordinates, where 2048 points are uniformly sampled from object surfaces.

\begin{figure}[ht]
  \centering
  \includegraphics[width=0.92\textwidth]{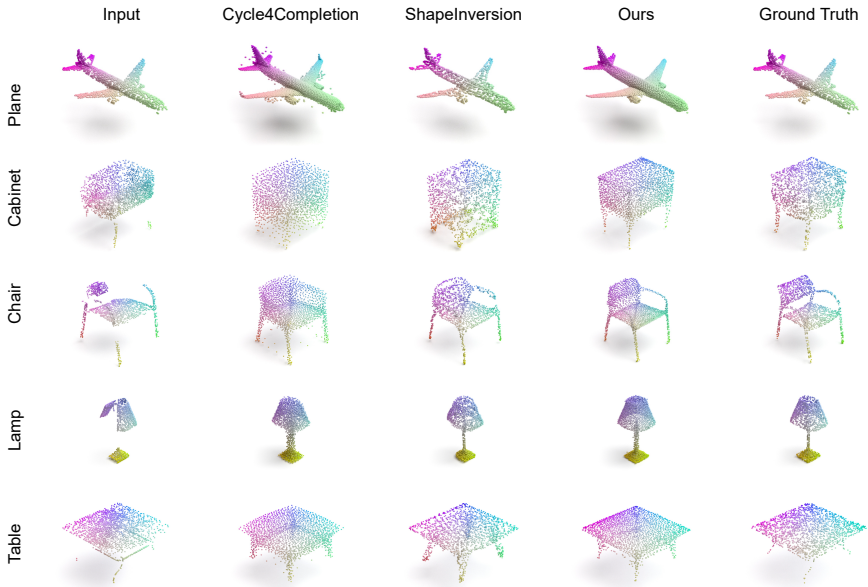}
  \caption{Qualitative Result on the 3D-EPN dataset. From left to right by column: input incomplete point clouds, results from Cycle4Completion~\cite{wen2021c4c} ShapeInversion~\cite{zhang2021unsupervised}, ours, and the ground truth.}
  \label{fig:qualitative}
\end{figure}

\noindent\textbf{Quantitative and qualitative evaluation}.
Table~\ref{tb:epn} shows performance comparison
% evaluation results
on the 3D-EPN dataset, where we adopt Chamfer Distance (CD) as the metric, and compare with both
% Previous
unsupervised methods~\cite{chen2020pcl2pcl, wen2021c4c, zhang2021unsupervised} and supervised methods~\cite{dai2017complete, 8578127, 8491026}.
% are compared together.
Table~\ref{tb:epn} explains
% It can be noticed
that our method outperforms existing unsupervised methods with a large margin.
Specifically, our method achieves the best result in all categories and surpasses Cai \emph{et al.}~\cite{Cai_2022_CVPR} by 4.2 on the average CD metric.
Figure~\ref{fig:qualitative} illustrates the qualitative results of our method compared with the competitive techniques.
In general, there are many outliers in the prediction of Cycle4Completion~\cite{wen2021c4c} while ShapeInversion~\cite{zhang2021unsupervised} struggles to yield shapes with uniformly distributed points.
It can be clearly observed that our method can achieve better quality with more uniform point distribution and complete geometries.

\begin{table}[tbp]
\centering
\caption{Shape Completion results of supervised (upper three rows) and unsupervised (lower five rows) methods on the 3D-EPN dataset. The numbers shown are CD
% Chamfer Distance (CD)
$\downarrow$ scaled by $10^4$.}
\addtolength{\tabcolsep}{-2pt}
\begin{tabular}{c|c|cccccccc}
\hline
Method           & Avg. & Plane & Cabinet & Car  & Chair & Lamp & Sofa & Table & Boat \\ \hline
3D-EPN~\cite{dai2017complete}           & 29.1    & 60.0  & 27.0    & 24.0 & 16.0  & 38.0 & 45.0 & 14.0  & 9.0  \\
FoldingNet~\cite{8578127}               & 9.2     & 2.4   & 8.5     & 7.2  & 10.3  & 14.1 & 9.1  & 13.6  & 8.8  \\
PCN~\cite{8491026}                      & 7.6     & 2.0   & 8.0     & 5.0  & 9.0   & 13.0 & 8.0  & 10.0  & 6.0  \\
% TopNet~\cite{8953650}                   & 8.4     & 2.5   & 8.8     & 5.9  & 9.3   & 12.0 & 8.4  & 13.5  & 7.1  \\ 
\hline
Pcl2Pcl~\cite{chen2020pcl2pcl}          & 17.4    & 4.0   & 19.0    & 10.0 & 20.0  & 23.0 & 26.0 & 26.0  & 11.0 \\
C4C.~\cite{wen2021c4c}                  & 14.3    & 3.7   & 12.6    & 8.1  & 14.6  & 18.2 & 26.2 & 22.5  & 8.7  \\
ShapeInv.~\cite{zhang2021unsupervised}   & 23.6     & 4.3   &  20.7   & 11.9 & 20.6  & 25.9 & 54.8 & 38.0   & 12.8  \\
Cai \emph{et al.} ~\cite{Cai_2022_CVPR}             & 13.6     & 3.5   &  12.2   & 9.0 & 12.1  & 17.6 & 26.0 & 19.8   & 8.6  \\
Ours                                    & \textbf{9.4}    & \textbf{2.3}   & \textbf{12.2}    & \textbf{5.8}  & \textbf{12.0}  & \textbf{12.8} & \textbf{10.3} & \textbf{13.8}  & \textbf{5.7}  \\ \hline
\end{tabular}
\label{tb:epn}
\end{table}

\subsection{Performance on Real-World Datasets}
To test the generalization ability of our model on  real-world data, we extract partial objects from MatterPort3D~\cite{Matterport3D}, ScanNet~\cite{dai2017scannet}, and KITTI~\cite{Geiger2013IJRR}. Our model, as well as other competitive state-of-the-art unsupervised methods, are trained on the 3D-EPN dataset without further fine-tuning.
As Table~\ref{tb:scan} indicates, our model can outperform Cycle4Completion~\cite{wen2021c4c} by a significant margin on all datasets.
However, since ShapeInversion~\citep{zhang2021unsupervised} directly minimizes the UCD metric between partial and prediction pairs, the results are comparable to ours on real-world datasets.

\begin{table}[tbp]
\centering
\caption{Shape completion on real-world scans. The numbers shown are Unidirectional Chamfer Distance (UCD)$\downarrow$ scaled by $10^4$. The \emph{sup} indicates supervised or unsupervised method.}
\begin{tabular}{l|l|ll|ll|l}
\hline
\multicolumn{1}{c|}{\multirow{2}{*}{Method}} & \multirow{2}{*}{sup.} & \multicolumn{2}{l|}{ScanNet}       & \multicolumn{2}{l|}{MatterPort3D}  & KITTI \\ \cline{3-7} 
\multicolumn{1}{c|}{}                        &                       & \multicolumn{1}{l|}{Chair} & Table & \multicolumn{1}{l|}{Chair} & Table & Car   \\ \hline
GRNet~\cite{xie2020grnet}                    & yes                   & \multicolumn{1}{l|}{1.6}   & 1.6   & \multicolumn{1}{l|}{1.6}   & 1.5   & 2.2   \\
PoinTr~\cite{Yu_2021_ICCV}                   & yes                   & \multicolumn{1}{l|}{1.7}   & 1.5   & \multicolumn{1}{l|}{1.8}   & 1.3   & 1.9   \\ \hline
C4C.~\cite{wen2021c4c}                       & no                    & \multicolumn{1}{l|}{12.0}  & 31.0  & \multicolumn{1}{l|}{12.0}  & 34.0  & 13.7  \\
ShapeInv.~\cite{zhang2021unsupervised}        & no                    & \multicolumn{1}{l|}{5.0}   & 3.0   & \multicolumn{1}{l|}{5.0}   & 3.0   & 6.0   \\
% *Stru.~\cite{Cai_2022_CVPR}                  & no                    & \multicolumn{1}{l|}{3.2}   & 2.7   & \multicolumn{1}{l|}{3.3}   & 2.7   & 4.2   \\
Ours                                         & no                    & \multicolumn{1}{l|}{4.0}   & 3.0   & \multicolumn{1}{l|}{4.0}   & 3.0   & 7.3   \\ \hline
\end{tabular}
\addtolength{\tabcolsep}{-2pt}
\label{tb:scan}
\end{table}

\subsection{Model Analysis}
\begin{figure}[tbp]
  \centering
  \includegraphics[width=0.92\textwidth]{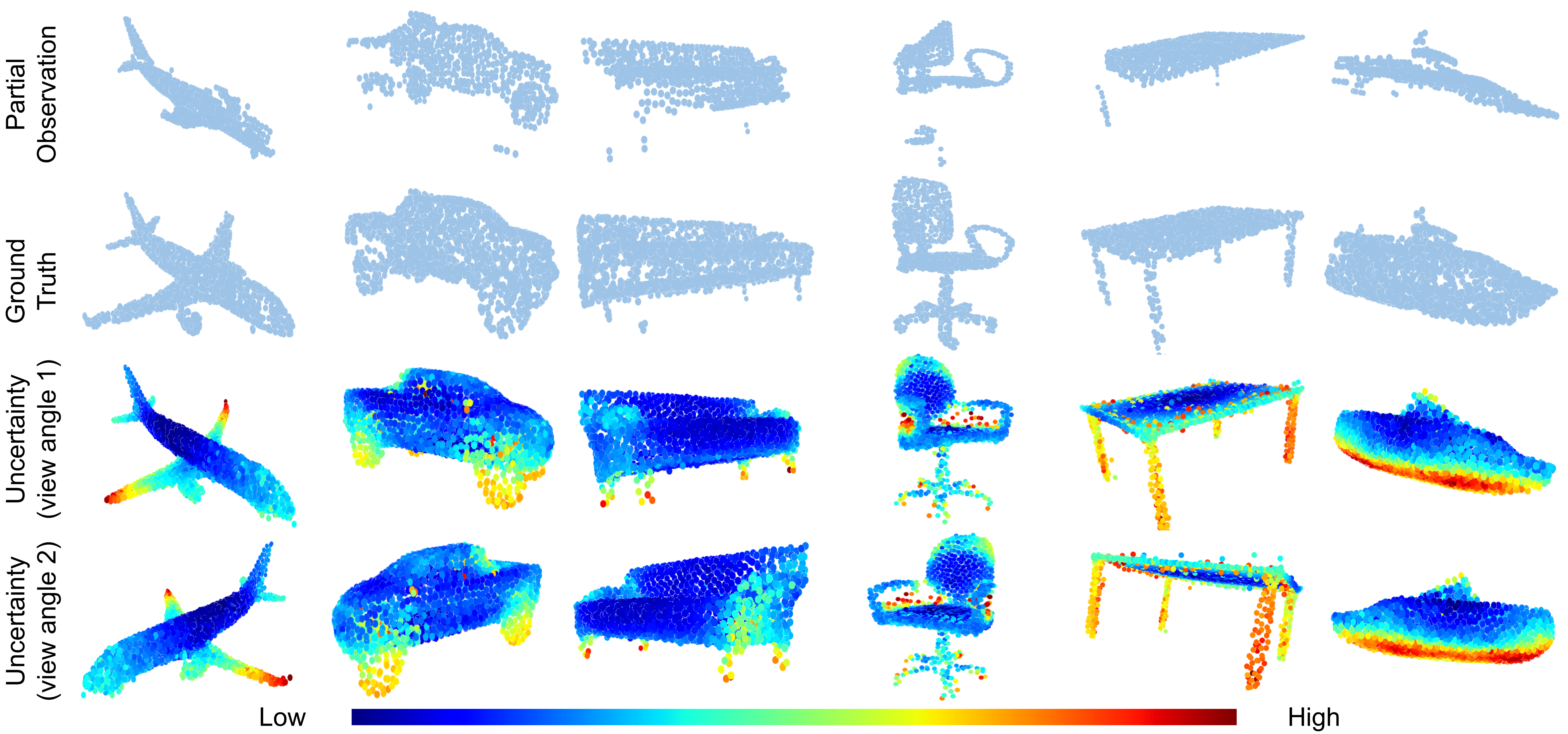}
  \caption{Uncertainty maps. The first and second rows show partial observations and their complete ground truth, and the third and fourth rows show two views of uncertainty maps.}
  \label{fig:conf}
\end{figure}

\noindent\textbf{Uncertainty Estimation:}
As the first generative unsupervised completion model, our framework can provide meaningful uncertainty maps that summarize the stochasticity of the inference process.
Specifically, an uncertainty map is generated by performing multiple times of inferences, and we define the variance of the multiple reconstructed complete point clouds as uncertainty following~\citep{what_kendall}.
% and computing the mean and variance per point.
Figure~\ref{fig:conf} visualizes sample uncertainty maps by rendering the per-point mean as a point and colorizing points with a heatmap converted from the magnitude of variance.
% where our uncertainty maps highlight the less accurate predictions, serving as tools to explain our predictions.
% are consistent with human perception.
The general trend is as follows:
1) an area with low uncertainty indicates that this region or its symmetric correspondence has been observed in the partial shape so that our model is confident with its prediction;
2) unobserved regions tend to have higher uncertainty since there can be multiple variants to form a valid completion.
For example, the plane wing is not observed in the partial shape, so our model assigns higher uncertainty scores when approaching the end of the wing.
Note that our model produces high uncertainty on outliers,
% prediction,
such as the abnormal points in the table or chair sample.

\noindent \textbf{Effect of Main Modules:}
To comprehensively analyse the proposed framework, we provide ablation studies to show the effectiveness of our proposed methods.
There are mainly three components that we deployed to the encoder-decoder backbone, \emph{i.e.}, energy-based latent transport, residual sampling, and adversarial loss.
We study their effect by removing one of them from the full model.
Table~\ref{tb:ebm} shows the full model's performance and performance of models with one of the above modules removed on the 3D-EPN~\cite{dai2017complete} dataset.

For each of the models with one module removed, we observe deteriorated performance on the four test categories, demonstrating the effectiveness of each module.
Note that the first row of Table~\ref{tb:ebm} indicates the capability of the encoder-decoder backbone.
The backbone is capable of predicting complete shapes with designed losses as well as the adversarial training framework.
The energy-based latent transport module boosts the performance by 1.4 in terms of the average CD.
However, suppose the model is not trained with the residual sampling strategy, in that case, the latent-space EBM can even harm the performance, as shown in the second row of Table~\ref{tb:ebm}.
The model without adversarial loss performs the worst since the adversarial loss regularizes predictions to be the complete shapes,
and if it is removed, the model tends to reconstruct the input (partial shapes) instead of completing the geometries.

\begin{table}[tbp]
\centering
\caption{Effect of different modules. CD$\downarrow$ scaled by $10^4$ are reported here}
\begin{tabular}{l|llll|l}
\hline
Method                & Chair         & Lamp          & Sofa          & Table         & Avg.          \\ \hline
w/o EB transport      & 13.6          & 14.3          & 11.9          & 14.7          & 13.6          \\
w/o residual sampling & 14.4          & 14.6          & 12.2          & 16.6          & 14.5          \\
w/o adversarial loss  & 16.6          & 19.5          & 17.7          & 25.5          & 19.8 \\
Full Model            & \textbf{12.0} & \textbf{12.8} & \textbf{10.3} & \textbf{13.8} & \textbf{12.2} \\  \hline
\end{tabular}
\addtolength{\tabcolsep}{-2pt}
\label{tb:ebm}
\end{table}

\section{Conclusions}
\label{sec:conclusion}

We present the first generative framework for unsupervised point cloud completion that is capable of generating prediction uncertainty maps.
Specifically, we introduce a latent code transport module based on a conditional EBM formulation for partial-to-complete transform.
A novel residual sampling strategy is proposed to facilitate end-to-end training for a latent-space EBM.
Our experiments show that the proposed method outperforms existing state-of-the-art models on both synthetic and real-world datasets.

%-------------------------------------------------------------------------

\bibliography{bmvc_final}
\end{document}